\documentclass[twoside,11pt]{article}

\usepackage{blindtext}

\usepackage{jmlr2e}

\usepackage{lastpage}
\usepackage{booktabs}
\usepackage{lipsum}
\usepackage{graphicx}
\usepackage{subcaption}

\ShortHeadings{}{}
\firstpageno{1}

\begin{document}

\title{Modalities, a PyTorch-native Framework For Large-scale LLM Training and Research}

\author{%
\name Max Lübbering$^{1,4}$ \email Max.Luebbering@iais.fraunhofer.de \\
\name Timm Ruland$^{1}$ \email Timm.Heine.Ruland@iais.fraunhofer.de \\
\name Richard Rutmann$^{1}$ \email richard.rutmann@iais.fraunhofer.de \\
\name Felix Stollenwerk$^{2}$ \email felix.stollenwerk@ai.se \\
\name David Fitzek$^{1}$ \email David.Fitzek@iais.fraunhofer.de \\
\name Michael Fromm$^{1,4}$ \email Michael.Fromm@iais.fraunhofer.de \\
\name Alexander Weber$^{1,4}$ \email Alexander.Weber@iais.fraunhofer.de \\
\name Rafet Sifa$^{1,3}$ \email Rafet.Sifa@bit.uni-bonn.de \\
\name Nicolas Flores-Herr$^{1}$ \email Nicolas.Flores-Herr@iais.fraunhofer.de\\
\name Joachim Köhler$^{1}$ \email Joachim.Koehler@iais.fraunhofer.de \\
\name Mehdi Ali$^{1,4}$ \email Mehdi.Ali@iais.fraunhofer.de \\
\addr $^{1}$Fraunhofer IAIS, Germany\\
$^{2}$AI Sweden, Sweden \\
$^{3}$University of Bonn, Germany \\
$^{4}$Lamarr Institute, Germany \\
}

\editor{}

\maketitle

\begin{abstract}
Today's LLM (pre-) training and research workflows typically allocate a significant amount of compute to large-scale ablation studies. Despite the substantial compute costs of these ablations, existing open-source frameworks provide limited tooling for these experiments, often forcing researchers to write their own wrappers and scripts. We propose Modalities, an end-to-end PyTorch-native framework that integrates data-driven LLM research with large-scale model training from two angles.
Firstly, by integrating state-of-the-art parallelization strategies, it enables both efficient pretraining and systematic ablations at trillion-token and billion-parameter scale. Secondly, Modalities adopts modular design with declarative, self-contained configuration, enabling reproducibility and extensibility levels that are difficult to achieve out-of-the-box with existing LLM training frameworks.

\end{abstract}

\begin{keywords}
  pretraining framework, distributed pretraining, large language models
\end{keywords}

\section{Introduction}
Due to the extreme costs of LLM pretraining, the margin for error during the ablation and pretraining phase is very small, requiring pretraining frameworks to exhibit a high level of maturity, extensive test coverage, and be thoroughly battle-tested. Given the high stakes and engineering complexity of LLM pretraining, most frameworks \citep{liang2024torchtitan, shoeybi2019megatron, harper2024nemo} provide a well-tested but limited set of model architectures and features. While these design decisions enable reliable training of LLMs in a fixed setting, they often come at the expense of extensibility, impairing LLM training in two ways. Firstly, for the final LLM training run, integrating emerging research insights or adapting to new training paradigms can introduce implementation and maintenance overhead, especially for external researchers and practitioners. Secondly, with LLM research and training being predominantly data-driven, the tooling must nevertheless allow for fast hypothesis testing and reproducible processing of a large amount of ablations at a trillion-token and billion-parameter scale. 

\begin{figure*}[t]
\centering
\includegraphics[width=1\textwidth]{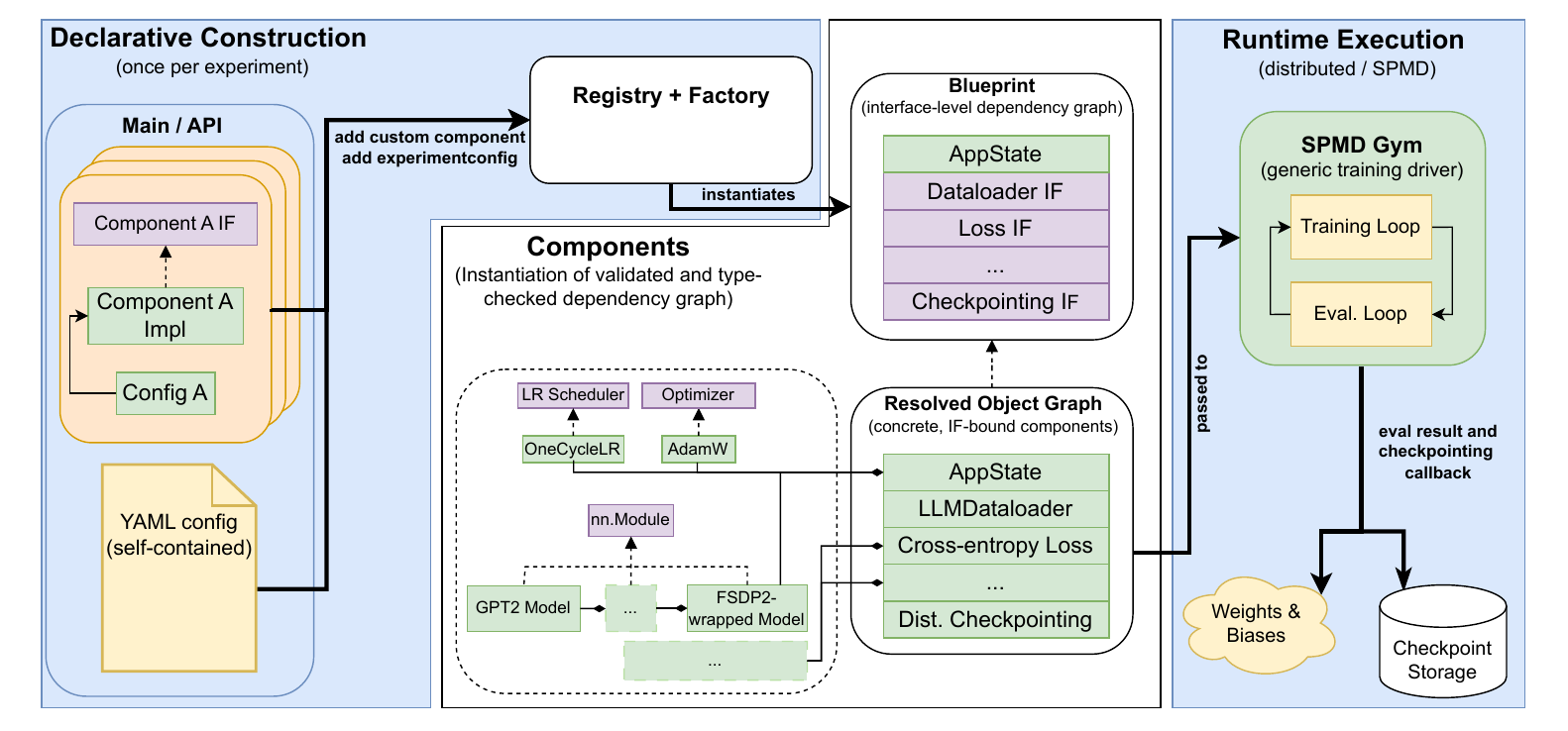}
\caption{High-level Modalities architecture. A self-contained YAML configuration defines an interface-level dependency graph that is resolved via a registry–factory mechanism into a resolved object graph. The resulting object graph is validated and injected into a generic SPMD training driver for distributed training and evaluation.}
\label{figure:architecture}
\end{figure*}

With Modalities\footnote{Modalities is open-sourced under MIT license: \url{https://github.com/Modalities/modalities}}, we set out to build a PyTorch-native LLM training framework that meets these two requirements. As an end-to-end solution, Modalities supports the full workflow from research-heavy data/architectural ablations towards production-ready LLM training (preprocessing, training, and downstream evaluation via Hugging Face compatibility). 
In contrast to Modalities, prominent LLM training frameworks (i.e., Nemo, Bridge, MegatronLM and TorchTitan) generally prioritize different workflows, focusing either on  production-oriented pre-training or research-driven experimentation. Nemo and Bridge provide complete training recipes for established architectures, following published configurations and primarily targeting practicioners who (pre-) train models along well-established, low-risk paths. 
TorchTitan and MegatronLM, by contrast, are more research-oriented and provide reduced reference implementations but hackable codebases as starting points for experimentation. While this design supports flexibility, it places the responsibility for integrating concrete ablation setups and experimental tooling into the existing codebases largely on the user.
Modalities avoids this trade-off by design. Its architecture enables research-driven experimentation and production-grade pre-training to coexist by expressing the full training setup as a declarative, self-contained object graph of pluggable components, thereby reducing integration effort while retaining scalability.

\section{Architectural Design to Unify Research and Pretraining}
\begin{figure}[t]
    \centering
    \begin{subfigure}[b]{0.32\textwidth}
        \centering
        \includegraphics[width=\linewidth]{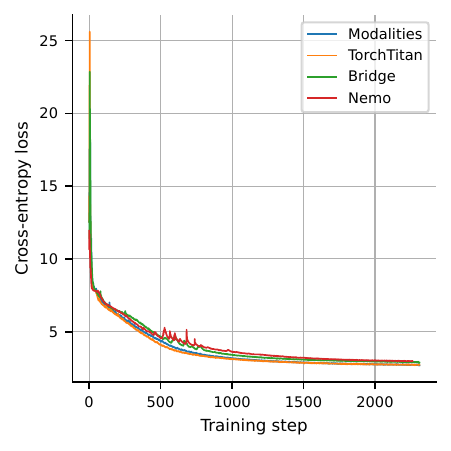}
        \caption{Training loss equivalence}
        \label{fig:convergenge}
    \end{subfigure}
    \begin{subfigure}[b]{0.32\textwidth}
        \centering
        \includegraphics[width=\linewidth]{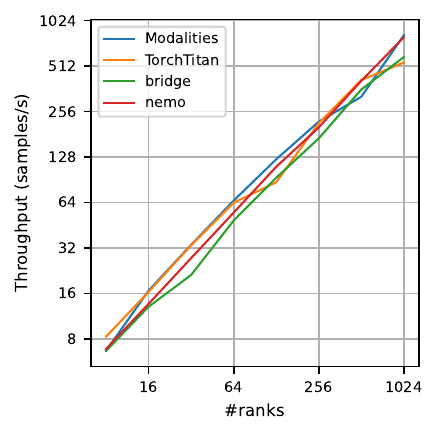}
        \caption{Training Throughput}
        \label{fig:scaling}
    \end{subfigure}
    \begin{subfigure}[b]{0.32\textwidth}
        \centering
        \includegraphics[width=\linewidth]{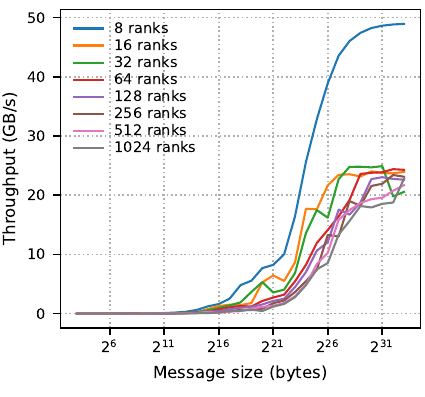}
        \caption{NCCL bandwidth}
        \label{fig:nccl}
    \end{subfigure}
    \hfill
    \caption{8B LLaMa 3 benchmarking on the Leonardo Supercomputer 
    \citep{turisini2023leonardo}: The left and center plots demonstrate equal convergence behavior on 100B Fineweb tokens \citep{penedo2024fineweb} and strong scaling behavior up to 1024 ranks. The standalone NCCL benchmark on the right shows latency/saturation behavior for different message size and different number of ranks on Leonardo.}
    \label{fig:benchmarks}
\end{figure}

Besides scalability and feature completeness for LLM training, the main design decisions of Modalities target extensibility and reproducibility, as shown in Fig.\,\ref{figure:architecture}. The user specifies the full training setup as a declarative, self-contained dependency graph within a YAML configuration file and can register custom components (e.g., a new model architecture) if required at runtime without having to adapt the framework. Note that Modalities already comes with 93 pluggable components each implementing one of the 32 pre-defined interfaces (IFs). Internally, Modalities instantiates the dependency graph leveraging a registry, factories and dependency injection, aggregates the top-level components within an object graph and passes the composition to the gym (inspired by \citet{lubbering2025architectural}). Misconfigurations are automatically flagged by the IF validation of the object graph, including custom components at runtime. Note that also other pipelines in Modalities, such as the tokenization or tracing pipeline, follow the same workflow but use different object graphs. 

By completely decoupling the experimental setup from the code and allowing for arbitrary custom components, Modalities offers a level of flexibility in ablation design that is essential for LLM training. To make these advantages concrete, consider ablating a new model architecture. In Modalities, this requires only implementing the model IF (\textit{nn.Module}) and registering the component via the registry API. Owing to the IF-level contract, the new architecture automatically integrates with existing components such as checkpointing and evaluation, and corresponding ablations can be specified entirely through the self-contained, declarative configuration. In existing frameworks, comparable extensions typically require forking the framework, integrating custom components into the training loop, and manually exposing new configuration paths.

\paragraph{Training Pipeline.}
Modalities applies the single-program multiple-data (SPMD) paradigm by leveraging the PyTorch-native parallelization strategy Fully Sharded Data Parallel (FSDP) \citep{10.14778/3611540.3611569}. 
While vanilla FSDP already scales up to 500 ranks \citep{liang2024torchtitan}, Modalities offers a combination with other parallelization strategies such as tensor parallelism, pipeline parallelism and hybrid sharded data parallel, allowing for scaling to 1000+ GPUs, as shown in Fig.\,\ref{fig:scaling}. Further, we implemented adaptable FSDP unit sizes, trading a slight memory overhead for improved NCCL bandwidth. Otherwise, with an increasing number of DP ranks, the small global NCCL message size (e.g., approx. 0.4\,MB per LLaMa 3 8B transformer block for DP-degree 1024) in the all-gather and reduce-scatter collectives do not saturate the interconnect anymore and become latency bound, as shown in Fig.\,\ref{fig:nccl}.

Looking beyond distributed training, Modalities exposes all major features and training regimes\footnote{For a full list see \newline \url{https://github.com/Modalities/modalities?tab=readme-ov-file\#supported-features}} as independently swappable components or configurable parameters (available across all training stages). To determine the optimal training setting for a given architecture and data mix, modalities provides hyperparameter search functionality for scalability / throughput optimization, and kernel / NCCL communication tracing.

\paragraph{Data Pipeline.}
Modalities provides efficient data processing routines including indexation, tokenization, chunking and global shuffling. Given a set of $n$ JSONL files, each file is processed in an massively parallel manner starting with indexation (identifying document boundaries), followed by tokenization. This results in $n$ memory-mapped files whose document index allows O(1) random access to tokenized documents.
Tokenization follows a producer-consumer design with a single reader and writer for document contiguous I/O, queues for batching and a configurable number of  tokenizer workers for parallel tokenization. This optimized design maintains achieves an average end-to-end throughput of 31M tokens per second which is 7x faster than the MegatronLM implementation\footnote{Benchmarked for tokenization throughput on a single DGX A100 with dual-socket AMD EPYC 7742 CPU (2 × 64 cores, 256 logical cores total) using a 100B-token subset of FineWeb-Edu \citep{penedo2024fineweb} and the Hugging Face LLaMA-3 tokenizer.}.

\paragraph{Integration.}
Modalities interfaces with the Hugging Face (HF) ecosystem in two ways, allowing for a straight-forward integration into most evaluation, inference and post-training frameworks. Firstly, any decoder-only model on HF is supported and can be trained or fine-tuned without any code changes required. Secondly, Modalities provides conversion routines to transform PyTorch-native (distributed) checkpoints into a HF-compatible format. 

\section{Conclusion \& Discussion}
We introduced Modalities, a PyTorch-native framework for distributed LLM training at scale. With its modular design, declarative configuration setup and seamless integration into the existing LLM ecosystem, Modalities offers extensibility and reproducibility while supporting 1000+ GPUs. Our architectural design enables researchers and practitioners to rapidly test out hypotheses without requiring any changes to the core code base.
\newpage
\acks{
This research has been funded by the Federal Ministry of Education and Research of Germany and the state of North-Rhine Westphalia as part of the Lamarr-Institute for Machine Learning and Artificial Intelligence.

We thank our collaborators from Fraunhofer IAIS (Alex Jude, Max Rudat, Sogol Haghighat, Julian Spravil, Thomas Dethmann, Santosh Thoduka, and David Berghaus), Fraunhofer IIS (Luzian Hahn, Viktor Haag, and Chris Hinze) and University of Bonn (Markus Frey, Behzad Shomali, and David Kaczer) for their open-source contributions to Modalities.
}

\vskip 0.2in
\bibliography{bib}

\end{document}